\documentclass[fleqn,10pt,x11names]{wlscirep_MR}
\usepackage[font=scriptsize]{caption} 

\usepackage[superscript,biblabel]{cite}
\usepackage{comment}
\usepackage[document]{ragged2e}
\usepackage[utf8]{inputenc}
\usepackage{multicol}
\usepackage{amssymb}
\usepackage{hyperref}
\usepackage{amsmath}

\usepackage[]{graphics}
\graphicspath{ {./figs/} }
\usepackage[rightcaption]{sidecap}
\usepackage{wrapfig}
\usepackage[normalem]{ulem}
\usepackage{makecell}
\usepackage{csquotes}
\usepackage{soul}
\usepackage{floatrow}
\usepackage{blindtext}
\usepackage{lipsum}

\usepackage{booktabs}
\usepackage{pdflscape}
\usepackage{longtable}
\usepackage{amssymb}
\usepackage{pdfpages}
\usepackage{subcaption}

\usepackage{placeins}
\usepackage{titlesec}
\usepackage{caption}
\usepackage{amsthm}
\usepackage{amsmath}

\usepackage{graphicx}

\title{The curious case of the test set AUROC}
\author[1,2,*,$\dagger$]{Michael Roberts}
\author[$\dagger$]{Alon Hazan}
\author[1,3,$\dagger$]{S{\"{o}}ren Dittmer}
\author[2]{James H.F. Rudd}
\author[1]{Carola-Bibiane Sch{\"{o}}nlieb}
\affil[1]{Department of Applied Mathematics and Theoretical Physics, University of Cambridge, Cambridge, UK}
\affil[2]{Department of Medicine, University of Cambridge, Cambridge, UK}
\affil[3]{ZeTeM, University of Bremen, Bremen, Germany}
\affil[*]{corresponding author: mr808@cam.ac.uk}
\affil[$\dagger$]{these authors contributed equally}

\usepackage{xcolor}

\begin{document}

\maketitle

\noindent

\begin{multicols*}{3}

\begin{justify}

\noindent
Whilst the size and complexity of ML models have rapidly and significantly increased over the past decade, the methods for assessing their performance have not kept pace. In particular, among the many potential performance metrics, the ML community stubbornly continues to use (a) the area under the receiver operating characteristic curve (AUROC) for a validation and test cohort (distinct from training data) or (b) the sensitivity and specificity for the test data at an optimal threshold determined from the validation ROC. 

However, we argue that considering scores derived from the test ROC curve alone gives only a narrow insight into how a model performs and its ability to generalise. In prior work, researchers have discussed the overall advantages and disadvantages of the AUROC \cite{halligan2015disadvantages} or proposed dropping it altogether in favour of other scoring systems \cite{chicco2023matthews}. We don't seek to discuss the individual shortcomings of the AUROC (e.g. equal treatment of all threshold values and extrapolation required for `degenerate' distributions) \cite{halligan2015disadvantages} or disregard the AUROC as it is a staple for ML researchers. However, we do seek to highlight, in particular, its shortcomings for evaluating a model's generalisability and motivate the community to providing solutions, some examples of which we also present. A more complete understanding may be obtained by supplementing (a) and (b) and reporting some simple additional scores, examples of which we will outline, that give insight into the generalisability of the model. To allow for rapid adoption of these additional scores, we share them in a public codebase \cite{codebase}.

The discussion in this paper focusses on the ROC but the arguments apply equally to the precision recall curve.

\end{justify}

\columnbreak
\begin{minipage}[htb!]{\dimexpr2\columnwidth+\columnsep}
    \centering
    \begin{minipage}[t]{.57\linewidth} 
        {\centering\hspace{0.85in}\textbf{Validation Set}}\vspace{0.1in}
    \end{minipage}\hspace{-0.5in}\vrule%
    \begin{minipage}[t]{.42\linewidth} 
    {\centering\hspace{1.05in}\textbf{Test Set}}\vspace{0.1in}
    \end{minipage}\\ \hrule
    \begin{minipage}[t]{.15\linewidth} 
        \rotatebox{90}{\parbox{6.0cm}{\centering\hspace{0.05in} \textbf{Coloured ROC} \hspace{0.20in} \textbf{Distributions}}}  
    \end{minipage}\hspace{-0.5in}\vrule
    \begin{minipage}[t]{.42\linewidth} 
        \includegraphics[width=\linewidth]{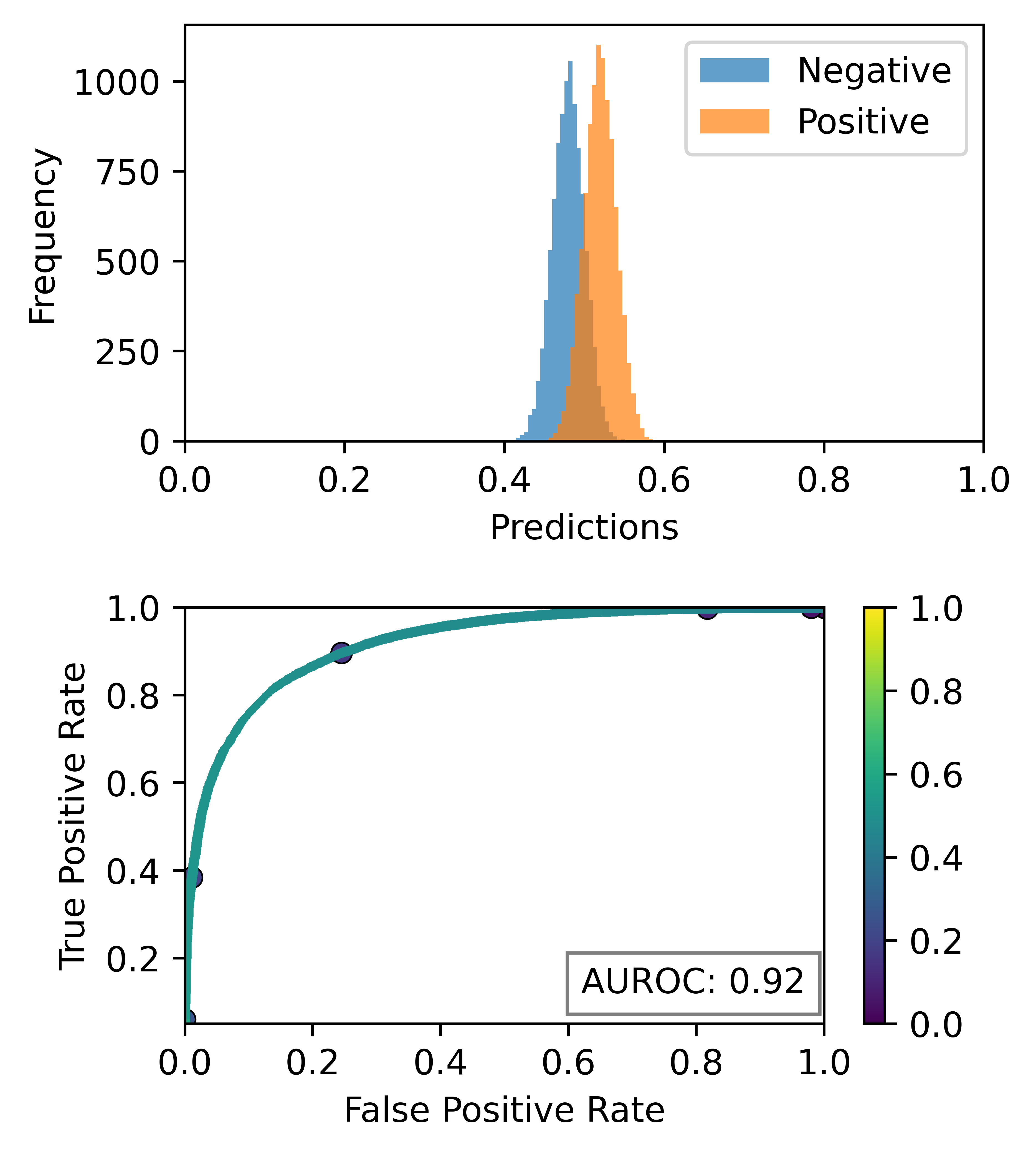}
    \end{minipage}\quad\vrule\quad%
    \begin{minipage}[t]{.42\linewidth} 
        \includegraphics[width=\linewidth]{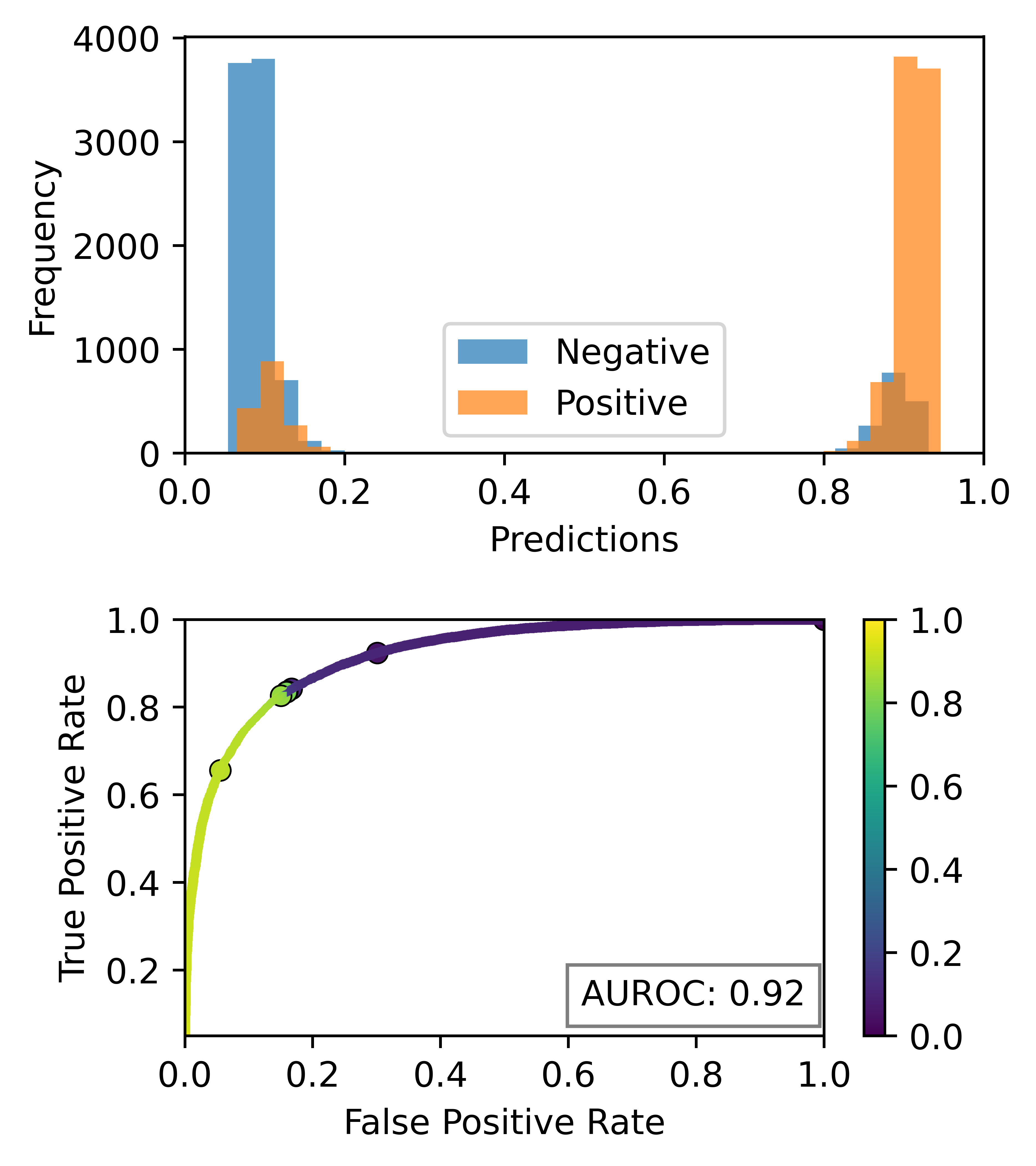}
    \end{minipage}
    \captionof{figure}{Example validation and test set model output distributions, ROC curves coloured by threshold.}\label{fig:one}
\end{minipage}\vspace{0.25in}\\

\begin{justify}

\noindent
\textbf{Some problems with the test AUROC}\\

\noindent
The key strength of the ROC curve is its simplicity. Simply by evaluating the sensitivity and specificity for model outputs at different thresholds, we gain great insight into how a model is performing. However, this simplicity is also a key limitation. We consider two key issues that are masked by the blunt usage of the ROC to assess model generalisation, namely distribution shift in the model outputs and erratic changes in sensitivity/specificity for small perturbations in the operating point.\\

\noindent
\textit{Distribution-shift in model outputs.} Suppose a model gives outputs $x\in\mathbb{R}^{n}$ for output dimension $n\in\mathbb{N}_{>0}$, then for any strictly monotonic function $f:\mathbb{R}\rightarrow\mathbb{R}$, the ROC curve will be identical for both $x$ and $f(x)$. Consequently, the AUROC, $\mathcal{A}$, for model outputs can be identical even when they have very different distributions.
\end{justify}

\columnbreak
\vspace*{17.5\baselineskip}

\begin{justify}
\noindent
In ML, it is quite common for a model to exhibit distribution shift in the model outputs when evaluated on datasets from different sources or different partitions of data from the same source. Therefore, it is possible to obtain consistently good AUROC values for a validation and test cohort of data whilst the outputs themselves exhibit significant distribution shift. See Figure~\ref{fig:one} for an example.

Practitioners often select a threshold as the model's operating point, where the ROC curve gives an optimal balance of sensitivity and specificity for their application. However, when distribution shift in the model outputs is present, this optimal threshold could be very different between the validation cohort and the test cohort. 

This gives rise to an overstatement in test set performance and compromises the utility of the model. In a real-world deployment scenario, when using an ML classification model on unseen external datasets, a threshold must be defined a-priori and applied to the model outputs to generate the test predictions. It is not sufficient to simply know that the validation and test set model outputs were concordant with one another.\\

\noindent
\textit{Optimal threshold sensitivity.} For an ML model to be of maximal utility, it is desirable for its reported sensitivity and specificity to be robust to a small change in the optimal threshold. However, this robustness is not considered in the AUROC computation and cannot be appreciated from the ROC curve. Two model outputs may have significantly different distributions, but if concordant, they will give identical ROC curves and AUROC values. See Figure~\ref{fig:one}.\\

\noindent
\textbf{A path to improving the test AUROC utility}\\

\noindent
Although we have focussed on the significant shortcomings in the use of the test AUROC in isolation to assess ML model performance, we believe it has some value. For ML practitioners, it provides a useful insight into the concordance of model outputs with the ground truth data for different datasets. However, we believe that this can be significantly enhanced by the reporting of additional scores which give insight into the underlying distributions of the model outputs. 

Overall, for a model with a ``good" test AUROC value, it is desirable that we are able to answer some additional questions. For example, for the distributions of the model outputs of the positive and negative classes: (a) how far are they from one another in the validation and test data? and (b) how robust are they to bias and noise? Additionally, for a particular threshold, how stable are the sensitivity and specificity values if applied to the validation and test data?

In the next section, we give examples of some scores which give answers to these questions. Importantly, these scores can all be evaluated from the existing model outputs,  requiring no additional experiments for the ML practitioner, and available in our codebase \cite{codebase}.\\


\columnbreak

\begin{minipage}[htb!]{\dimexpr2\columnwidth+\columnsep}
    \centering
    \begin{minipage}[t]{.57\linewidth} 
        {\centering\hspace{0.85in}\textbf{Validation Set}}\vspace{0.1in}
    \end{minipage}\hspace{-0.5in}\vrule%
    \begin{minipage}[t]{.42\linewidth} 
    {\centering\hspace{1.05in}\textbf{Test Set}}\vspace{0.1in}
    \end{minipage}\\ \hrule
    \begin{minipage}[t]{.15\linewidth} 
        \rotatebox{90}{\parbox{8.0cm}{\centering\textbf{Single dataset discrepancy scores}}}  
    \end{minipage}\hspace{-0.5in}\vrule
    \begin{minipage}[t]{.42\linewidth} 
        \includegraphics[width=\linewidth]{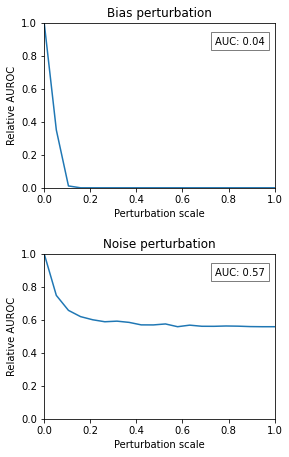}
    \end{minipage}\quad\vrule\quad%
    \begin{minipage}[t]{.42\linewidth} 
        \includegraphics[width=\linewidth]{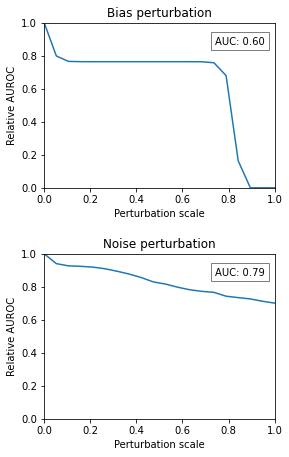}
    \end{minipage}
    \captionof{figure}{The robustness to bias and noise for the model outputs on the example validation and test data.}\label{fig:two}
\end{minipage}

\vspace*{1.35\baselineskip}

\noindent
\textbf{Potential solutions}\\

\noindent
There are two classes of scores which we will explore and present examples of. Firstly, there are those that can be computed for model outputs derived from a data cohort in isolation. These allow for a study of model outputs on validation data whilst the model is still being developed and for further improvements to be made. Secondly, we consider scores that compare model outputs between validation and test cohorts. These allow for some understanding of how the model outputs change and give insight into its ability to generalise.\\

\textbf{Single dataset discrepancy scores.} For simplicity, let us consider a binary classification ML model and a dataset whose samples generate predictions $\hat y\in\mathbb{R}^{n}$ which are paired with ground truth labels $y\in\{0,1\}^{n}$. 

\noindent
Samples that have $y_{i}=1$ will be termed the positive class and those with $y_{i}=0$ are the negative class. Our ambition is to understand how far the distributions of the model outputs for the positive and negative classes are from one another. We highlight just two potential solutions that measure: (a) robustness to bias and (b) robustness to noise.
Denoting the AUROC as $\mathcal{A}_y(\hat y)$ and a dis-



\end{justify}
\columnbreak

\vspace*{22.5\baselineskip}
\begin{justify}
tributional shift of strength $\sigma$ as $\hat y_\sigma = \mathcal{P}_y^\sigma(\hat y)$, we define the robustness to this type of distributional shift as
\[\frac{1}{\mathcal{A}_y(\hat y)}\int_{\sigma_{min}}^{\sigma_{max}} \mathcal{A}_y(\hat y_\sigma) d\sigma.\]

\noindent
\textit{Robustness to bias.} This is a measure of how much bias must be added to the negative or positive class for it to have minimal distribution distance from the other. To do this, we define $\mathcal{P}_y^\sigma(\hat y)$ via $\hat y_\sigma|_{y=0} = \hat y|_{y=0}$ and $\hat y_\sigma|_{y=1} = y|_{y=1} - \sigma$, where the subtraction is read to be pointwise. See Figure~\ref{fig:two}.\\

\noindent
\textit{Robustness to noise.} If the distributions of model outputs for the positive and negative classes are close to one another, see Figure~\ref{fig:one}, the reported model performance is highly variable, dependent on careful selection of the choice of threshold. To quantitatively assess this, we diffuse $\hat y$ by increasing its variance using $\mathcal{P}_y^\sigma(\hat y) \sim \mathcal{N}(\hat y, \sigma^2I)$. See Figure~\ref{fig:two}.\\

\noindent
\textbf{Multiple data cohort discrepancies.} If we have the model outputs for both the validation and test cohorts, we can start to consider how performance drifts between them. We highlight two examples that focus on assessing the drifts (a) in the sensitivity and specificity at different thresholds (b) the 2-Wasserstein ``earth-mover" distances between the the positive and negative classes in the validation and test set.\\

\noindent
\textit{Drift in sensitivity and specificity.} For a fixed threshold applied to the model outputs for the validation and test cohorts, $y^{(V)}$ and $y^{(T)}$ respectively, we determine how much the model's sensitivity and specificity drift. This can be assessed using any distance metric; for simplicity, we focus on the Euclidean distance. Integrating these distances over all thresholds gives us the average drift score $\mathcal{D}_{\text{sens/spec}}$. Formally, 
\[
\mathcal{D}_{\text{sens/spec}} = \int_{0}^{1} \left| \mathcal{S}(\hat y^{(V)},\tau) - \mathcal{S}(\hat y^{(T)},\tau) \right|^{2}_{2} \mathrm{d}\tau
\]
where $\mathcal{S}_y(\hat y,\tau)\in[0,1]^{2}$ returns the sensitivity and specificity of the model outputs $\hat y$ at threshold $\tau$. See Figure~\ref{fig:three}. \\

\noindent
\textit{Wasserstein distance matrix.}
The objective is two-fold: firstly, we want the distance between $\hat y^{(V)}|{y^{(V)} = 0}$ and $\hat y^{(T)}|{y^{(T)} = 1}$, as well as between $\hat y^{(T)}|{y^{(T)} = 0}$ and $\hat y^{(T)}|{y^{(T)} = 1}$ to be large. This ensures clear demarcation and less ambiguity between the positive and negative classes within each data cohort.

Concurrently, we want the distance between $\hat y^{(V)}|{y^{(V)} = 0}$ and $\hat y^{(T)}|{y^{(T)} = 0}$, and also between $\hat y^{(V)}|{y^{(V)} = 1}$ and $\hat y^{(T)}|{y^{(T)} = 1}$ to be small. This fosters alignment and consistency within the same categories across the two datasets.

In essence, the goal is to promote distinction between different classes while ensuring coherence within them. Many different distance measures can be used to achieve this; we choose the 2-Wasserstein ``earth-mover" distances. We may conveniently represent these relationships with a distance matrix, as in Figure~\ref{fig:four}, where we want large entries on the diagonals and small entries otherwise.\\

\noindent
\textbf{Next steps}\\

\noindent
This manuscript is intended to serve as a ``call to arms" for the community of ML practitioners that there are severe deficiencies in the current approaches for evaluating model performance. We have highlighted several potential methods that allow for assessing distribution shifts and drifts in performance

 \columnbreak 
 
 between validation and test data. However, these are not exhaustive and are not necessarily the best approaches. There needs to be a systematic, robust and concerted effort to consider these issues by the ML community and we hope that this paper serves as a starting point for such discussion and research.

\typeout{}
\bibliography{references}


\begin{minipage}{\dimexpr2\columnwidth+\columnsep}
    \centering
        \begin{minipage}[t]{\linewidth} 
        {\centering\hspace{1.3in}\textbf{Drift scores between multiple cohorts}}\vspace{0.1in}
        \end{minipage}\\ \hrule
    \begin{minipage}[t]{\linewidth} 
            \includegraphics[width=0.8\linewidth]{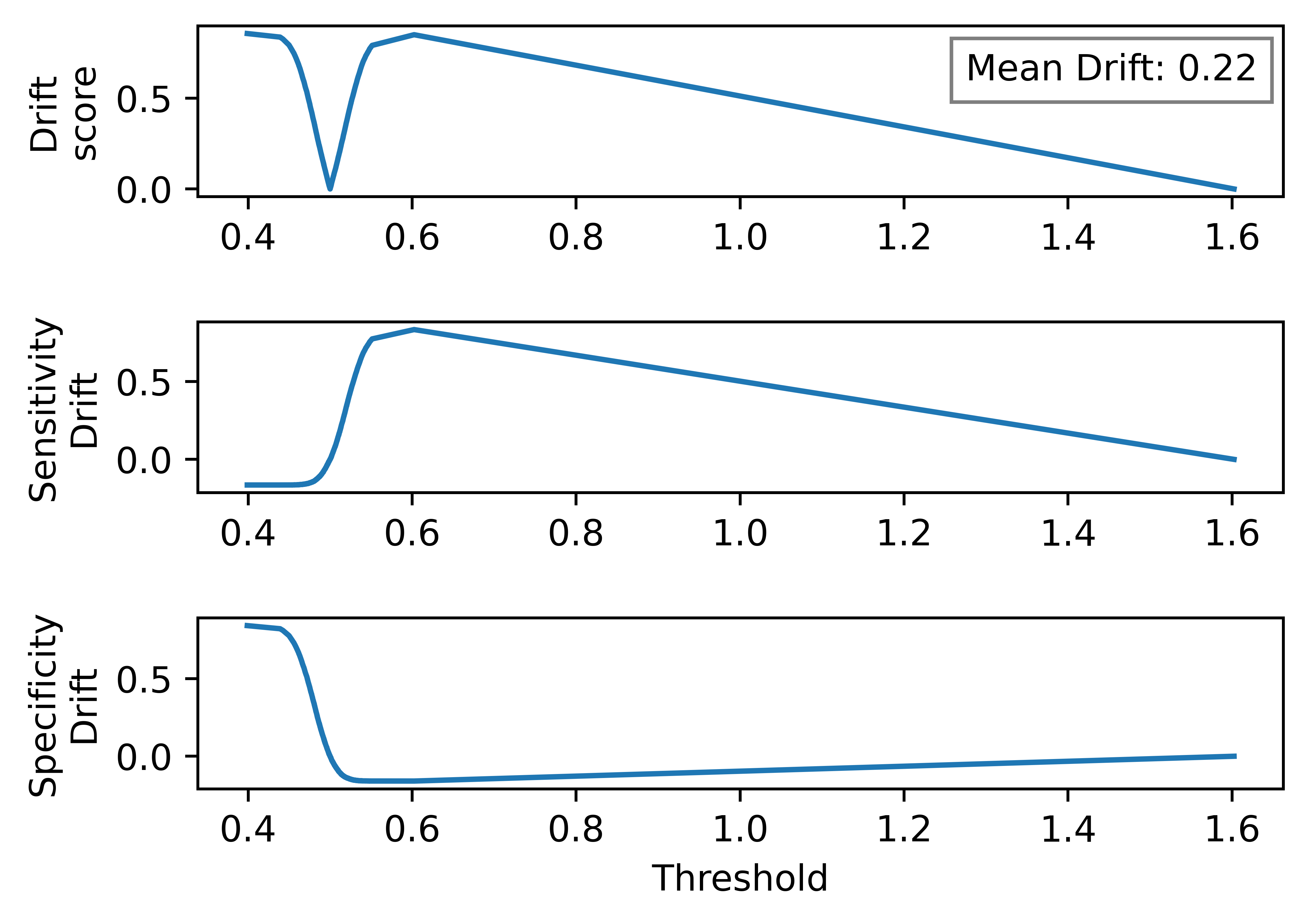}
    \end{minipage}
    \captionof{figure}{The overall drift scores between the example validation and test cohorts (top) with the sensitivity and specificity disaggregated (middle and bottom, respectively).}\label{fig:three}
\end{minipage}\\ \vspace{0.4in}

\begin{minipage}{\dimexpr2\columnwidth+\columnsep}
    \centering
    \begin{minipage}[t]{0.5\linewidth}
       \centering
       \begin{tabular}{l|rr}
           {} &    ${y^{(V)} = 1}$ &    ${y^{(T)} = 0}$ \\
           \midrule
           ${y^{(V)} = 0}$ &  0.04 &  0.38 \\
           ${y^{(T)} = 1}$ &  0.38 &  0.56 \\
       \end{tabular}
    \end{minipage}%
    \captionof{figure}{The Wasserstein distance matrix, which shows the transport costs between the positive and negative classes of the validation and test set.}
    \label{fig:four}
\end{minipage}


\end{justify}

\end{multicols*}

\end{document}